# A Novel Approach for Dimensionality Reduction and Classification of Hyperspectral Images based on Normalized Synergy

Asma Elmaizi[1*], Hasna Nhaila[2], Elkebir Sarhrouni[3], Ahmed Hammouch[4], Nacir Chafik[5]
Research Laboratory in Electrical Engineering LRGE,
Mohammed V University, Rabat, Morocco

*Abstract*—During the last decade, hyperspectral images have attracted increasing interest from researchers worldwide. They provide more detailed information about an observed area and allow an accurate target detection and precise discrimination of objects compared to classical RGB and multispectral images. Despite the great potentialities of hyperspectral technology, the analysis and exploitation of the large volume data remain a challenging task. The existence of irrelevant redundant and noisy images decreases the classification accuracy. As a result, dimensionality reduction is a mandatory step in order to select a minimal and effective images subset. In this paper, a new filter approach normalized mutual synergy (NMS) is proposed in order to detect relevant bands that are complementary in the class prediction better than the original hyperspectral cube data. The algorithm consists of two steps: images selection through normalized synergy information and pixel classification. The proposed approach measures the discriminative power of the selected bands based on a combination of their maximal normalized synergic information, minimum redundancy and maximal mutual information with the ground truth. A comparative study using the support vector machine (SVM) and k-nearest neighbor (KNN) classifiers is conducted to evaluate the proposed approach compared to the state of art band selection methods. Experimental results on three benchmark hyperspectral images proposed by the NASA "Aviris Indiana Pine", "Salinas" and "Pavia University" demonstrated the robustness, effectiveness and the discriminative power of the proposed approach over the literature approaches.

## I. INTRODUCTION

In the next decade, the exploitation of hyperspectral imaging [1] will experience a spectacular development thanks to the technological imaging evolution growing in many areas. The current generation of hyperspectral sensors provides large quantities of precise information on the nature and spatial-temporal evolution of the analyzed areas. The maturity and accessibility of this technology make it possible to address new applications in the fields of agronomy, environment, military, industrial and health security, etc. In remote sensing [2], the rich and detailed spectral information provided by hyperspectral images helped in detecting the composition of imaged materials and classifying targets with high spectral and spatial accuracy [3]. Embedded on an aircraft, a hyperspectral sensor operating in the visible near-infrared range (400-1000 nm) can simultaneously record several tens, even hundreds of narrow spectral bands. The volumes of data (data cubes) acquired often reach gigabytes for a single scene observed. As a result, their exploitation with classical methods developed for monochrome or color is very limited. In many cases, it is unnecessary to process all the spectral bands of an HSI [4][5] (Hughes phenomenon).

Most materials have specific characteristics only at certain bands, which makes the remaining spectral bands somewhat redundant. Additionally, some noisy bands [6] are influenced by various atmospheric effects. To overcome these challenges and respond quickly to the needs arising from the different potential applications, dimensionality reduction is an essential pre-processing step. Methods of bands selection must be developed to achieve the best compromise between reducing and preserving the amount of information acquired.

The selection approaches [7] consist of retaining the dataset physical meaning by selecting the most relevant bands. The hyperspectral band's selection will be the main topic of the work presented in this paper. Currently, selection algorithms can be categorized into two common approaches: wrapper and filters [8].

- The wrapper methods are classifier-dependent. They evaluate the band's relevance based on the classification accuracy and generally reach promising results. However, these approaches are very expensive in terms of computational complexity and may suffer from over-fitting to the learning algorithm.

- The filter methods are classifier-independent. They are based on the maximization of a certain evaluation function. The main advantages of these methods are their computational efficiency, simplicity and independence from the classifier. A common drawback shared by these approaches in literature is the lack of information about the synergy and interaction correlation between the picked bands and the ground truth.

In literature, many filter-selection methods have been developed using different evaluation measures. The evaluation function is generally based on distance, information, correlation and different consistency measures.

*Corresponding Authors

Information theory introduced by "Cover & Thomas" [9] has been widely applied in filter methods, where information measures are used to evaluate the band's relevance and quantify the amount of information contained on images. This paper contributes to the knowledge in the area of hyperspectral dimensionality reduction by proposing a new approach based on normalized synergic correlation. The proposed method aims to overcome the limitations of the current state of the art filter band selection methods such as overestimation of the band significance, which causes selection of redundant and irrelevant bands. The new evaluation method selects the band that has maximum relevance, minimum redundancy and maximum normalized synergy with the previously selected bands. This paper reviews the state of art band selection methods highlighting their common limitations and comparing their performance versus the proposed algorithm. Experimental results are carried out using three benchmark hyperspectral images proposed by the NASA "AVIRIS Indiana Pine" [10], "Pavia University" and "ROSIS Salinas" [11]. Classification results are generated using the SVM [12][13] and KNN [14] to demonstrate the effectiveness and classification accuracy improvement of the proposed approach.

The rest of the paper is structured as follows. Section 2 describes the fundamentals of information theory and reviews the state of art band selection methods. Section 3 presents the proposed normalized max synergy (NMS) algorithm. Section 4 outlines the experiment conducted on the three datasets and analysis the achieved results. Finally, Section 5 concludes the paper.

## II. Background on Information Theory based Approaches

In this section, we describe some basic concepts about information theory and feature selection, which will be used to build the proposed hyperspectral band selection algorithm.

The information theory proposed by "Cover & Thomas" [9] has been widely applied in filtering methods, where information measures are used to assess the relevance and discrimination of the characteristic.

Definition 1: The Shannon entropy introduced in (1) is defined as the quantification of the amount of information contained in variable X.

$$H(X) = \sum_X P(X) \log_2 P(X) dX \qquad (1)$$

Since, Shannon entropy H(X) is defined for a single variable and it is independent of the class, the mutual information between two random variables was introduced in order to measure the statistical dependence between the features and between the features and the class.

Definition 2: The mutual information (MI) of a pair of variables in (2) represents their degree of dependence in the probabilistic sense. It is the reduction of uncertainty on a random variable through the knowledge of another.

$$MI(X, Y) = \sum_{X,Y} P(X, Y) \log_2 \frac{P(X,Y)}{P(X)P(Y)} dXdY \qquad (2)$$

$$MI(X, Y) = H(X) + H(Y) - H(X, Y) \qquad (3)$$

$$MI(X, Y) = H(X) - H(X|Y) \qquad (4)$$

The P(X,Y) in (2) is the joint probability function and P(X), P(Y) represent the marginal probabilities.

In the equation (3), H(X) and H(Y) are the Shannon entropies of two variables X, Y respectively and H(X, Y) is the joint entropy between the variables. The mutual information can also be formulated using the conditional entropy as presented in (4).

Mutual information has the following properties.

- Mutual information is positive or zero.
- The mutual information is symmetrical.

In a wide survey of the feature selection literature, we have identified different information theory-based filters [15] and we will be presenting a selection of the most well-known criteria.

In the results section, the selected relevant methods will be applied to hyperspectral data to compare it with our proposed approach.

Battiti [16] proposed to use mutual information for variable selection in the Mutual Information-based Feature Selection (MIFS) algorithm. In this approach, the number of variables is fixed in advance and at each step, the variable that maximizes the mutual information between all the variables already selected is chosen. Formally, the variable selected by the MIFS algorithm is the one that maximizes the following goal function:

$$MIFS = Argmax(MI(Fi, C) - \beta \sum_{Fs \in S} MI(Fi, Fs)) \qquad (5)$$

The factor 'β' in (5) allows to control the redundancy term MI(Fi,Fs) and has a great influence on the selection algorithm. Several authors like Bollacker and Ghosh [17] that use different values for the parameter β without any justification. The value of β is often determined experimentally and depends on the data used. The problem is highlighted when the subset is very large and the redundancy term becomes larger than the relevance term. The algorithm will then select irrelevant features because they are not redundant, but not because they are relevant to the class.

As a consequence, several variants of the MIFS algorithm have been proposed in recent years in order to overcome its limitations. Kwak and Choi [18] proposed the algorithm MIFS-U as an improvement of MIFS.

$$MIFSU = Argmax(MI(Fi, C) - \beta \sum_{Fs \in S} \frac{MI(C,Fs)}{H(Fs)} MI(Fi, Fs)) \qquad (6)$$

Peng [19] analyzed as well the limitations of the previous selection approach and proposed a robust approach minimum redundancy maximum relevance (mRMR) where the redundancy term in (7) is divided over the cardinality of the subset.

$$mRMR = Argmax(MI(Fi, C) - \frac{1}{S} \sum_{Fs \in S} MI(Fi, Fs)) \qquad (7)$$

Asma et al. [20] proposed a hybrid strategy combining the filter mRMR with the Fanno based wrapper strategy in order to select the relevant hyperspectral bands. Yang and Moody [21] proposed the Joint Mutual Information (JMI) (10) based on maximizing the cumulative summation of Joint Mutual Information of the selected subset.

The joint mutual information is presented in (9):

$$I(X, Y|Z) = H(X|Z) - H(X|Z, Y) \quad (8)$$

$$JMI(X, Y; Z) = I(X, Z|Y) + MI(X, Y) \quad (9)$$

The JMI filter approach is defined in (10) as follow:

$$JMI = \text{Argmax} \sum_{Fs \in S} JMI(Fi, Fs; C) \quad (10)$$

Meyer and al. introduced the Double Input Symmetrical Relevance (DISR) based on the joint mutual information as well [22]. The goal function of this approach is based on the symmetrical relevance as illustrated in (11).

$$DISR = \text{Argmax}\left(\sum_{Bs \in S} \frac{I(Fi, Fs; C)}{H(Fi, Fs; C)}\right) \quad (11)$$

Within information theory studies dedicated to hyperspectral images selection, GUO [23] proposed an effective MI-based filter algorithm to select the most discriminative bands. He calculates the average of bands 170 to 210 of the HSI AVIRIS 92AV3C image that will be introduced in the result part in order to produce an estimated ground truth map (class), and use it instead of the real ground truth. Sarhrouni [24] proposed a mutual information based filter approach (MIBF) considering that the band maximizing the mutual information with the class or ground truth is a good approximation of it and introduced a threshold "Th" in order to control the redundancy criteria.

To select the most relevant bands, Nhaila presented also an enhanced version of the normalized mutual information [25]. This approach uses normalized MI to control the redundancy instead of MI as defined in the equation. This method is reported to perform well in terms of classification accuracy and stability.

### III. Proposed Filter Approach based on Maximum Normalized Synergy MNS

#### A. Limitation of State of Art Methods

According to the previous part of this article, there are two factors that affect bands selection: MI between the bands and the ground truth (relevancy term) and MI between the selected bands (redundancy term). A remarkable weakness is that the majority of the feature selection algorithms did not consider the synergic dependence between the candidates bands with the other bands already selected. The methods discussed previously work on the assumption that the relevance of a single band is associated with the degree of dependence of this band on the ground truth. But it may happen that some HSI bands acting independently do not provide any additional information to the classification but when grouped together with other images gives promising results.

#### B. The Proposed Approach Normalized Mutual Synergy (NMS)

Aiming at the shortcomings of the above algorithms, a new band selection algorithm is proposed as an enhancement of the state of art methods. The proposed approach considers three factors: relevancy, redundancy and normalized synergy in order to select the discriminative bands. The purpose of the proposed approach is to reformulate the band selection problem as a modelling problem based on multi-criteria. We formalize this multi-criteria into three types of interaction between hyperspectral bands.

- Relevance criteria

The decision of which bands are the most relevant and should be selected is usually associated with the degree of dependency of each single band to the ground truth (class). This criterion is calculated using the mutual information between the selected band and the ground truth MI(Bi,GT) and it is reflecting the shared information between the selected band Bi and the ground truth and will be used to evaluate the discriminative ability of each band to the classification.

- Redundancy criteria

This criterion is a reflection of the common information shared by the selected bands. The amount of redundancy can never decrease when other new bands are added. The redundancy will be controlled using the normalized mutual synergy NMS that will be presented in the next part.

- Complementary criteria

The decision of which bands are the most discriminative and powerful for target classification is usually associated with the degree of complementarity and synergy between the selected bands. The complementarity will be evaluated using the normalized interaction information NMS. This measure will be used to control simultaneously the redundancy and the complementarity between the picked bands.

In fact, Guo [23] and Sarhouni [24] uses MI(Fi,Fs) to measure the bands redundancy. They considered that all correlated bands are redundant but neglect that some of the wrongly removed bands are synergic. The normalized mutual information algorithm (NMI) [25] was proposed as an enhancement of mutual info based methods. According to the NMI algorithm, there are two kinds of bands correlations: Independent correlation when the measure is 0 and redundant correlation when the measure is between 0 and 1. This approach considers all the correlated bands as redundant and would wrongly judge the synergic bands as redundant as well. To overcome the limitations discussed previously, we propose the NMS algorithm that can provide a more accurate measure for band interaction including the three criteria (relevance, redundancy and complementarity).

Definition 3: The interaction information I(X;Y;Z) or synergy S(X,Y) has been defined by Jakulin [26] as the decrease in uncertainty caused by joining the attributes X and Y in a Cartesian product [27]. Considering the bands and the class label simultaneously, the bands synergy in (12) can be defined as follows:

$$S(X, Y) = I(X, Y, Z) = I(X, Y) - I(X, Y|Z) \quad (12)$$

Substituting eq. (3) for I(X,Y) and I((X,Y)|Z) in (13)

$$(I(X, Y, Z) = [H(X) + H(Y) - H(X, Y)] - [H(X|Z) + H(Y|Z) - H((X, Y)|Z] \quad (13)$$

From (13) we can deduce the relationship between mutual information and the synergy measure as below:

$$S(X, Y) = I(X, Y, Z) = I\big((X, Y), Z\big) - I(X, Z) - I(Y, Z) \quad (14)$$

The normalized mutual synergy is defined in (15) as follow:

$$NMS(X, Y, Z) = \frac{2\, I(X,Y,Z)}{I(X,Z)+I(Y,Z)} \quad (15)$$

**Proposed Algorithm Normalized mutual synergy NMS**

**Inputs :**

- Hyperspectral Dataset H
- S={b1,b2,...,bn} Spectral bands of the HSI image
- GT: the class label or ground truth that will be used for supervised classification
- A defined percentage of pixels for training and pixels for testing.

**Outputs:**

- The subset R of k selected bands ranked in the selection order.
- Reproduced ground truth.
- Classification results and metrics.

1. Set R ← [ ] "empty result set in the initialization step"
2. Select the relevant band that got the max mutual information with the class label (ground truth)

$$b^* = Argmax\ b_i \in S\ (MI\ (b_i, GT))$$

Set S←S-{b*}, R←{b*}

3. Calculate the ground truth estimated using the first selected band

Set GTest=b*

4. While |R|≤ k during each iteration choose the band that maximizes the objective function:

$$F(b^*) = Argmax\big(MI(b_i, GT) + NMS(b_i, GTest, GT)\big)$$

$$F(b^*) = Argmax\Big(MI(b_i, GT) + 2 * \frac{I(b_i, GTest, GT)}{MI(b_i, GT) + MI(GTest, GT)}\Big)$$

5. Recalculate the class label estimated using the new selected band in each iteration:

$$GTest = \frac{GTest + b*}{2}$$

S←S-{b*}, R←R∪{b*}

6. Output the result subset R containing the selected bands.
7. Evaluate the selected bands using a classifier function.
8. Generate the classification metrics and the reproduced ground truth.

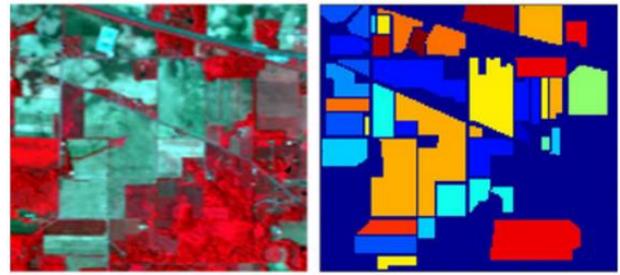

Fig. 1. Proposed Normalized Synergy Algorithm (NMS).

In the proposed NMS filter approach, the normalized mutual synergy criteria will be used to evaluate the redundancy and complementarity between bands as follow:

- The NMS measure is positive: 0<NMS≤1 when the selected bands have synergic correlation and together provide critical info for accurate classification which cannot be provided by each one of them individually.

- The NMS measure is negative: -1≤NMS<0 when the selected bands are redundantly correlated and provide redundant information that does not help to increase the classification accuracy.

- The NMS measure is equal to zero when the selected band is independent from the already selected bands in the context of the class label of ground truth.

Let S={b1,b2,...,bn} the band's set of hyperspectral image dataset H with n bands. The goal is to find a subset of bands that maximizes the objective function f(x) based on the NMS measure and which represents the combination of these three types of interaction. In order to avoid testing all possible combinations that will cause a computational burden, we propose a greedy selection algorithm (Fig. 1) that begins with an empty set of bands. The first chosen band will be based on the relevance criteria and thus will be the one with the maximum mutual information with the ground truth and successively adds bands during each iteration that maximize the objective function f(x) combining the three criteria. Afterward, the selected band's classification rate will be evaluated using the classifiers Support Vector Machine (SVM) and K-Nearest (KNN).

## IV. EXPERIMENTAL RESULTS AND ANALYSIS

In order to evaluate the performance of the proposed filter method, we will use three real hyperspectral datasets from NASA's Airborne Visible Infra-Red Imaging Spectrometer (AVIRIS) [10] and the Reflective Optics System Imaging Spectrometer (ROSIS) [11]. These images are one of the most challenging classification problems since they are overlaid with mixed pixels and similar classes and they will be presented shortly on this section. Then, we will introduce as well the classifiers and evaluation metrics that will be used for results analysis. Finally, the experimental results for each hyperspectral image are presented and discussed by comparing with the close literature methods.

## A. Experimental Datasets

*1) Aviris indiana pines:* The Indiana pines [10] is an agricultural image acquired over the Indian Pines test site in North-western Indiana, USA and collected by the Airborne Visible Infra-Red Imaging Spectrometer. The Aviris Indiana pines hyperspectral image is built using a 3-dimensional cube with two spatial and a third spectral as illustrated in Fig. 2.

Every material and compound of the earth surface illustrated in the ground truth (Fig. 3) is identified with its unique electromagnetic signature. It consists of 224 spectral reflectance bands in the wavelength range 0.4–2.45 μm covering 145*145 pixels. This scene contains two-thirds agriculture (alfalfa, corn, oats, soybean, wheat), and one-third forest (woods, and different sub-classes of grass or other natural perennial vegetation). The ground-reference data of the scene is designated into 16 classes with a total of 10,366 labelled samples (Fig. 3).

*2) Rosis pavia university:* The Pavia scene was recorded using ROSIS (Reflective Optics System Imaging Spectrometer) [11] over the Pavia University in Italy. It consists of 103 spectral reflectance bands in the wavelength range 0.43- 0.86 μm covering 610*340 pixels. This scene covers an urban environment that is mainly constituted of Natural objects (trees, meadows, soil), various solid structures (asphalt, gravel, metal sheets, bricks) and shadows. The ground-reference data of the scene is designated into 9 classes as illustrated in Fig. 4.

*3) Aviris salinas:* The Salinas scene was acquired over the Salinas Valley in California and gathered by the AVIRIS sensor [10] as well. The scene consists of 224 spectral reflectance bands covering $512 \times 217$ pixels and it mainly contains covering vegetables, fields and bare soils. The ground-reference image of the scene is designated into 16 classes and presented in Fig. 5.

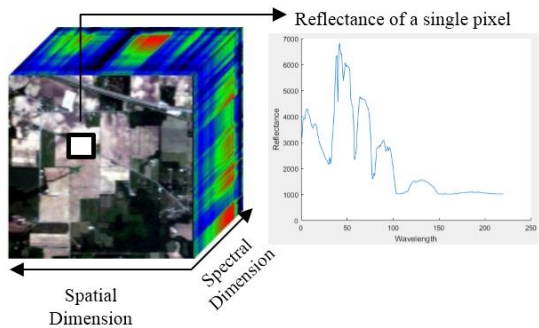

Fig. 2. Data Cube of the Hyperspectral, Image Aviris Indiana Pines.

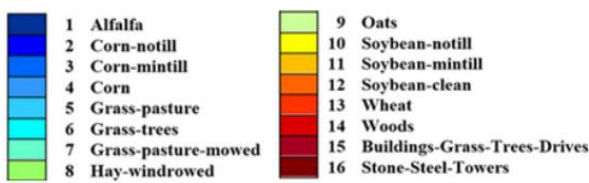

Fig. 3. (Right) Ground Truth Data of the Indian Pines image. (Left) Three-Band Color Composite of the Indian Pines Image (Bands 30, 43, and 21).

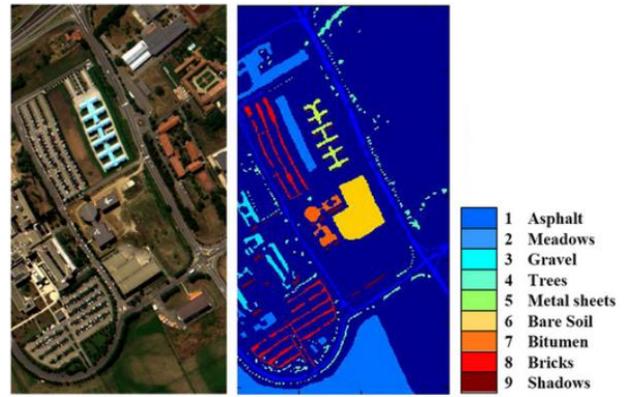

Fig. 4. (Right) Ground Truth Data of the Pavia University Image. (Left) Three-Band Color Composite of the Indian Pines Image (Bands 13, 33, and 56).

## B. Classifiers and Evaluation Metrics

To assess the performance of the proposed method, the support vector machine (SVM) [12][13] and the k-nearest neighbor (KNN) [14] were chosen for the classification step. The Gaussian radial basis function (RBF) kernel is adopted for the SVM classifiers and the cross-validation operation is processed in order to determine the optimal parameters C and γ of the RBF kernel. The KNN algorithm is used with the Euclidean distance and k=3 nearest neighbors.

The SVM and KNN algorithms choice is based on our previous comparison study [28] where both classifiers results showed their great performance for HSI classification compared to other classifiers.

In all experiments, 10%, 25% and 50% of instances in each class are randomly labelled to compose the training sets and the remaining pixels are considered for the test and validation.

In order to evaluate the performance of the proposed method and compare it with other literature approaches, we calculate several well-known metrics in the literature reflecting the classification accuracy performance (OA, AA and KC).

- The Overall accuracy (OA) refers to the number of correctly classified instances divided by the total number of testing samples.

- The Average accuracy (AA) is a measure of the mean value of the classification accuracies of all classes.

- The kappa coefficient (KC) is a statistical measurement of consistency between the ground truth map and the final classification map.

## C. Classification Results and Discussion

*1) Classification results on HS image aviris indiana pine:* Table I presents a comparison of classification results between the proposed approach (NMS) versus the information theory-based filters (MIBF, JMI, DISR and NMI). We carry out a set of parallel experiments in order to calculate the overall accuracy, the average accuracy and the kappa coefficient for the selected bands. Each column of Table I represents the 16 individual class accuracies of the Indiana scene using the SVM and the KNN classifiers. The results reflect the robustness and strength of the proposed approach in selecting highly discriminative bands. In all cases, classification accuracies decreased when using the KNN instead of the SVM for the classification stage.

This result confirms the fact that the classifier SVMs are less affected by the Hughes phenomenon especially when trained with mixed spectral-spatial data confirming the results obtained on our previous work [28]. Fig. 6 presents the classification accuracy rate of the proposed approach with regard to the other band selection algorithms for different selected bands number up to 80 selected bands. From the Indiana pines results, we remarked that the MIBF algorithm has the lowest classification accuracy rate due to the weak band's correlation estimation. The redundancy term in this method is affected by the choice of the threshold Th and is often determined experimentally based on the dataset used. Additionally, this algorithm performance decreases when the subset is very large and the redundancy term becomes larger than the relevance term. The JMI and DISR algorithms provide promising results compared to MIBF and NMI due to the joint mutual info based objective function that gives better bands estimation.

The JMI reaches 91.49 % for 60 selected bands which is more than NMI by 2.86% and MIBF 2.32%. Our proposed approach outperform the other methods with (OA=94,09% & Kappa =93.69%, AA=94.3%) for 40 selected bands.

Experimental results reflected the effect of the normalized synergy adopted in the objective function of our algorithm. In fact, the correlation between bands is evaluated and classified into three types of interaction: relevancy, redundancy and synergic. It is worth noting also that the NMS proposed algorithm selects the high discriminative bands with a high speed as shown in Fig. 6 for small selected bands number.

During each iteration, the band is only selected when it increases the objective function based on the three correlation types.

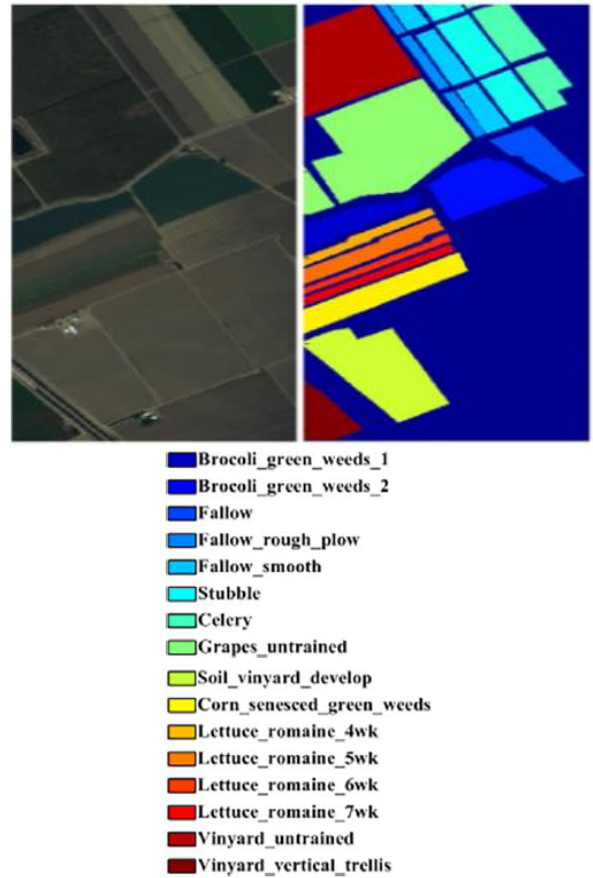

Fig. 5. (Right) Ground Truth Data of Salinas Image. (Left) Three-Band Color Composite of the Salinas Image (Bands 10, 30, and 21).

TABLE. I. CLASSIFICATION RESULTS OF THE PROPOSED APPROACH VERSUS different LITERATURE APPROACHES IN INDIANA PINES DATASET AND USING KNN & SVM (40 SELECTED BANDS)

| Algorithm Indiana | MIBF | | NMI | | JMI | | DISR | | NMS (Proposed approach) | |
|---|---|---|---|---|---|---|---|---|---|---|
| Classifier | KNN | SVM | KNN | SVM | KNN | SVM | KNN | SVM | KNN | SVM |
| 1 | 68,52 | 81,48 | 55,56 | 74,07 | 64,81 | 87,04 | 66,67 | 87,04 | 83,33 | **90,74** |
| 2 | 76,43 | 75,24 | 73,85 | 70,08 | 74,55 | 71,2 | 76,78 | 74,06 | 77,75 | **91,63** |
| 3 | 78,9 | 80,22 | 72,3 | 71,82 | 66,31 | 66,19 | 70,5 | 78,18 | 76,14 | **89,93** |
| 4 | 61,54 | 75,64 | 60,26 | 76,92 | 58,12 | 77,78 | 66,24 | 84,19 | 73,5 | **89,32** |
| 5 | 84,91 | 89,13 | 83,1 | 83,5 | 94,16 | 97,38 | 94,57 | 98,19 | 94,57 | **98,79** |
| 6 | 95,05 | 94,91 | 95,18 | 92,24 | 98,39 | 98,39 | 98,13 | 98,53 | 97,46 | **98,53** |
| 7 | 42,31 | 76,92 | 26,92 | 69,23 | 50 | 84,62 | 65,38 | 80,77 | 80,77 | **92,31** |
| 8 | 96,32 | 97,75 | 95,71 | 95,91 | 96,93 | 98,57 | 97,55 | 98,36 | 98,57 | **99,18** |
| 9 | 55 | 95 | 85 | 90 | 80 | 90 | 80 | 90 | 80 | **100** |
| 10 | 70,97 | 72,52 | 73,97 | 74,07 | 67,15 | 54,55 | 70,56 | 64,67 | 88,12 | **92,46** |
| 11 | 83,71 | 85,74 | 83,1 | 85,25 | 84,48 | 86,39 | 85,49 | 87,52 | 88,7 | **93,64** |
| 12 | 76,06 | 88,27 | 71,5 | 82,9 | 68,57 | 74,59 | 70,52 | 75,57 | 73,62 | **92,35** |
| 13 | 97,17 | 98,11 | 95,28 | 95,75 | 99,06 | 99,06 | 98,58 | 99,53 | 98,11 | **98,11** |
| 14 | 94,44 | 94,36 | 91,73 | 94,82 | 96,14 | 98,15 | 96,75 | 98,3 | 96,45 | **98,38** |
| 15 | 52,89 | 59,21 | 51,58 | 54,74 | 62,37 | 81,84 | 63,16 | 81,58 | 57,89 | **86,58** |
| 16 | 86,32 | 96,84 | 82,11 | 96,84 | 84,21 | 94,74 | 86,32 | 95,79 | 93,68 | **96,84** |
| Kappa | **80,71** | **83,31** | **78,93** | **80,73** | **80,05** | **81,23** | **81,87** | **84,26** | **85,38** | **93,69** |
| AA | **76,28** | **85,08** | **74,82** | **81,76** | **77,83** | **85,03** | **80,45** | **87,02** | **84,92** | **94,3** |
| OA | **81,92** | **84,35** | **80,25** | **81,93** | **81,3** | **82,4** | **83** | **85,24** | **86,29** | **94,09** |

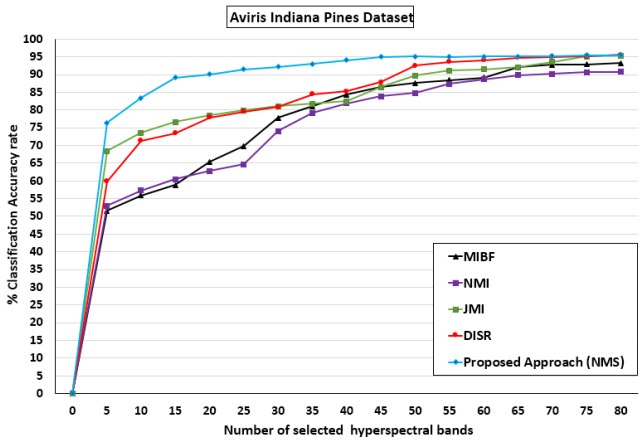

Fig. 6. The Classification Accuracy Rate Results Versus the Number of Selected Bands using the SVM Classifier for the Indiana Pines Image.

Fig. 7 presents reproduced classification maps using 40 selected bands using the proposed algorithm (NMS). The obtained result confirms that the selected 40 bands are highly discriminative to distinguish and classify the scene materials and target.

The classification accuracy reaches 94.09 % using the support vector machine classifier and we notice that 40 bands selected using the NMS approach are sufficient to detect all scene materials and provide a produced maps close to the ground truth.

*2) Classification results on HS image rosis pavia:* Table II illustrates the classification accuracy rate of the proposed algorithm compared to four literature approaches using SVM and KNN classifiers. Each row of the table provide the individual accuracies of the Pavia scene and the last three rows generate the overall, average and kappa classification metrics. From the produced result, we confirm the robustness of the proposed approach that outperforms the other mutual information-based filters for the Pavia scene. In fact, the normalized synergy method selects relevant bands rapidly due to the accurate objective function used during the selection process for both classifiers SVM and KNN. Our proposed method achieves an overall accuracy of 94.7% classification accuracy for 40 selected bands, which is higher than the JMI by 5.04% and DISR by 4.03. Fig. 8 present the evaluation of the proposed approach with regard to other features selection algorithm defined in the literature for the Pavia scene. The analysis of the results of the different curves confirms that the evaluation of bands correlation using the NMS helps in improving the results significantly compared with other algorithms. Fig. 9 illustrates the reproduced maps using 40 selected bands by the NMS. The dimensionality reduction of the pavia scene into 40 pertinent bands allows classification of image pixels and detection of material and target of the scene with high accuracy. The selected bands are enough to discriminate the material in 9 classes of the image and reproduce a close map to the ground truth.

*3) Classification results on HS image aviris salina:* The following Table III and Fig. 10 confirms the robustness of our proposed method that outperforms the other filters for this dataset as well. It is remarkable that the JMI and DISR algorithms in a second-place perform better than the other filters based on mutual information. This result is due to the joint mutual information objective function that provides an accurate bands evaluation. Using the KNN classifier, we achieved OA=92.56%, kappa=92.06% and AA=96.27% for 40 bands.

TABLE. II. CLASSIFICATION RESULTS OF THE PROPOSED APPROACH VERSUS DIFFERENT LITERATURE APPROACHES IN PAVIA UNIVERSITY DATASET AND USING KNN & SVM (40 SELECTED BANDS)

| Algorithm | MIBF | | NMI | | JMI | | DISR | | NMS (Proposed approach) | |
| --- | --- | --- | --- | --- | --- | --- | --- | --- | --- | --- |
| Classifier | KNN | SVM | KNN | SVM | KNN | SVM | KNN | SVM | KNN | SVM |
| 1 | 89,02 | 91,86 | 91,83 | 93,08 | 91,92 | 93,68 | 92,02 | 94,33 | 92,99 | **94,06** |
| 2 | 97,66 | 97,69 | 96,49 | 96,9 | 96,66 | 97,06 | 96,95 | 97,11 | 98,34 | **98,34** |
| 3 | 71,18 | 67,98 | 75,75 | 76,08 | 75,66 | 69,22 | 76,04 | 70,75 | 77,61 | **77,7** |
| 4 | 86,06 | 95,59 | 89,36 | 92,72 | 89,3 | 90,7 | 89,52 | 90,86 | 90,57 | **96,44** |
| 5 | 99,63 | 100 | 99,48 | 100 | 99,48 | 99,85 | 99,48 | 99,93 | 99,63 | **100** |
| 6 | 78,72 | 83,4 | 68,7 | 71,76 | 67,05 | 62 | 68,28 | 67,77 | 84,39 | **92,24** |
| 7 | 86,84 | 74,66 | 86,54 | 84,66 | 86,02 | 82,33 | 86,84 | 85,34 | 90,68 | **81,88** |
| 8 | 86,58 | 86,12 | 89,76 | 89,63 | 87,02 | 89,87 | 86,77 | 90,11 | 90,33 | **90,36** |
| 9 | 100 | 100 | 99,89 | 99,89 | 99,68 | 99,68 | 99,68 | 99,68 | 99,89 | **99,79** |
| Kappa | 89,64 | 90,9 | 89,04 | 90,09 | 88,61 | 88,37 | 88,99 | 89,5 | 92,63 | **94,04** |
| AA | 88,41 | 88,59 | 88,65 | 89,41 | 88,09 | 87,16 | 88,4 | 88,43 | 91,6 | **92,31** |
| OA | 90,79 | 91,91 | 90,25 | 91,19 | 89,88 | 89,66 | 90,21 | 90,67 | 93,44 | **94,7** |

TABLE. III. CLASSIFICATION RESULTS OF THE PROPOSED APPROACH VERSUS DIFFERENT LITERATURE APPROACHES IN SALINAS DATASET AND USING KNN AND SVM (40 SELECTED BANDS)

| Algorithm Salinas | MIBF | | NMI | | JMI | | DISR | | NMS (Proposed approach) | |
|---|---|---|---|---|---|---|---|---|---|---|
| Classifier | KNN | SVM | KNN | SVM | KNN | SVM | KNN | SVM | KNN | SVM |
| 1 | 95,17 | 94,47 | 98,86 | 96,62 | 98,46 | 97,46 | 99,9 | 98,56 | **99,35** | 99,25 |
| 2 | 87,47 | 92,3 | 99,84 | 99,89 | 98,15 | 97,56 | 99,97 | 99,97 | **99,89** | 100 |
| 3 | 99,7 | 99,7 | 98,53 | 97,27 | 98,84 | 98,73 | 99,19 | 98,48 | **99,49** | 99,24 |
| 4 | 98,92 | 96,34 | 99,71 | 99,64 | 99,57 | 99,5 | 99,71 | 99,5 | **99,64** | 99,5 |
| 5 | 98,43 | 98,39 | 99,1 | 97,61 | 97,91 | 96,71 | 99,03 | 96,79 | **99,33** | 98,81 |
| 6 | 99,55 | 99,07 | 99,97 | 99,97 | 99,55 | 99,12 | 99,95 | 99,97 | **99,95** | 99,95 |
| 7 | 87,96 | 88,13 | 99,86 | 99,72 | 99,53 | 99,75 | 99,86 | 99,75 | **99,44** | 99,8 |
| 8 | 83,46 | 88,87 | 83,67 | 89,1 | 87,65 | 89,73 | 84,09 | 91,89 | **85,25** | 88,97 |
| 9 | 99,27 | 99,27 | 99,65 | 99,27 | 99,53 | 99,44 | 99,74 | 99,24 | **99,56** | 99,32 |
| 10 | 94,02 | 91,21 | 95,55 | 90,82 | 95,61 | 89,99 | 95,33 | 91,58 | **95,76** | 90,58 |
| 11 | 97,28 | 95,04 | 96,82 | 90,17 | 94,01 | 89,42 | 97,1 | 91,57 | **98,31** | 94,3 |
| 12 | 99,12 | 98,91 | 99,27 | 98,6 | 99,48 | 98,34 | 99,69 | 99,07 | **99,79** | 96,63 |
| 13 | 97,82 | 98,47 | 99,45 | 99,02 | 98,58 | 98,58 | 99,24 | 99,13 | **98,69** | 99,69 |
| 14 | 94,58 | 91,4 | 97,2 | 93,93 | 96,45 | 95,14 | 97,94 | 93,46 | **95,14** | 98,03 |
| 15 | 66,24 | 45,2 | 70,07 | 41,99 | 79,8 | 51,51 | 70,45 | 37,42 | **72,36** | 93,36 |
| 16 | 95,57 | 94,85 | 98,89 | 94,58 | 98,84 | 98,78 | 99,39 | 96,79 | **98,28** | 53,74 |
| Kappa | 88,49 | 86,59 | 91,37 | 87,59 | 93,3 | 89 | 91,64 | 87,8 | **92,06** | 98,23 |
| AA | 93,41 | 91,98 | 96,03 | 93,01 | 96,37 | 93,74 | 96,29 | 93,32 | **96,27** | 94,93 |
| OA | 89,21 | 87,43 | 91,91 | 88,36 | 93,72 | 89,69 | 92,16 | 88,56 | **92,56** | 90,62 |

Fig. 11 illustrates the reproduced ground truth for 40 bands using the SVM Classifier. Using the bands selected by our approach, we were able to detect the 16 classes included in the Salinas scene with an accuracy equal to 90.62.

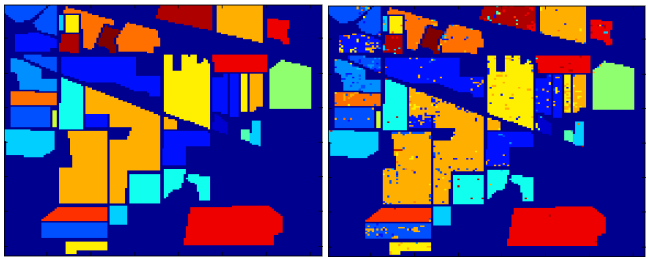

Fig. 7. Indiana Pines Ground Truth (Left), Indiana Ground truth Reproduced Map (Right) using the Proposed Algorithm NMS for 40 Bands (94.09%).

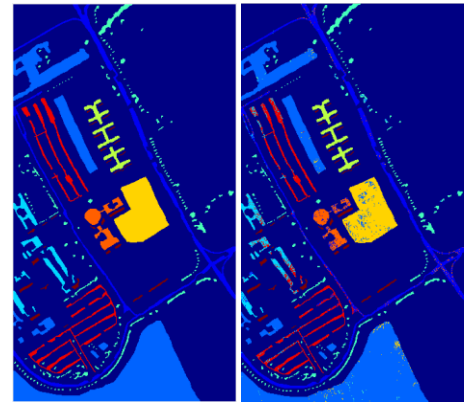

Fig. 9. University Pavia Ground Truth (Left), Pavia Reproduced Map (Right) using the Proposed Algorithm NMS for 40 Bands (OA=94.7%).

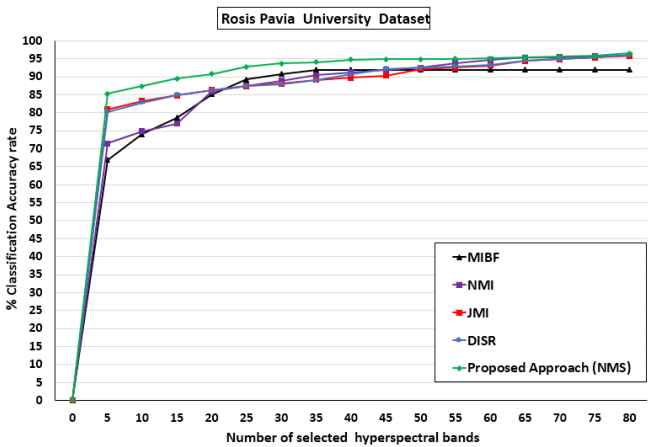

Fig. 8. The Classification Accuracy Rate Results Versus the Number of Selected Bands using the SVM Classifier for the Pavia University Image.

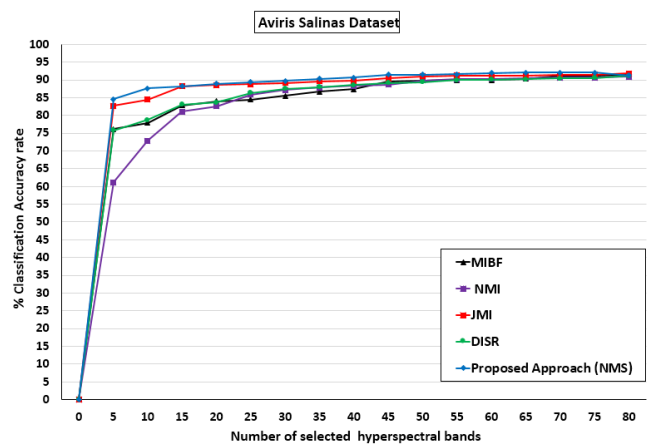

Fig. 10. The Classification Accuracy Rate Results Versus the Number of Selected Bands using the SVM Classifier for the Salinas Image.

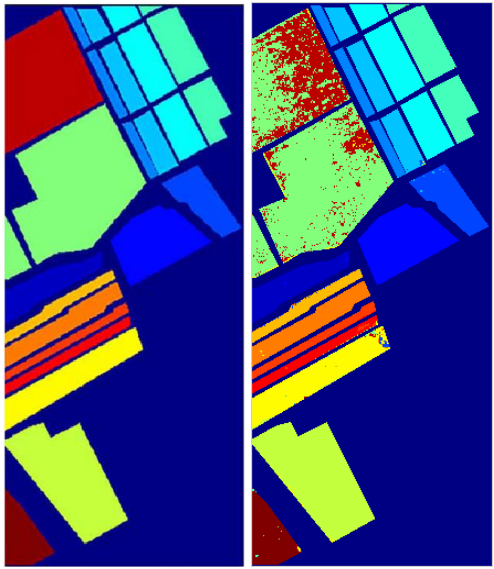

Fig. 11. Salinas Ground Truth (Left), Salinas Reproduced Map (Right) using the Proposed Algorithm NMS for 40 Bands (OA=90.62%).

## V. Conclusion

In the last decades, remote sensing community has achieved a great improvement in detecting targets and classifying materials of difficult scenes due to the hyperspectral technology. However, the high dimensionality reduction of this type of images had always been a necessity in order to detect materials with high classification accuracy. In this paper, we present a new band selection approach based on information theory: normalized mutual synergy. This method is designed in order to resolve the problem of the high dimensionality of the hyperspectral images by selecting the discriminative bands, removing redundant and noisy ones. This method is based on the evaluation of every single band of the hyperspectral cube based on an objective function maximization. The evaluation function is a combination between the three correlation types: normalized synergy, redundancy and relevancy.

The robustness and effectiveness of the proposed approach have been evaluated using three hyperspectral public datasets by the NASA. Experimental results using the SVM and KNN classifiers confirm that the proposed approach increases the classification accuracy significantly and helps in selecting high discriminative bands rapidly. Compared to the other filter methods, our algorithm evaluates the band's correlation and interaction with high accuracy in order to select the discriminative bands and helps in detecting the scene materials to provide a reproduced map close to the ground truth.

Future work includes more experiments using other hyperspectral datasets and including new spectral parameters in order to improve bands evaluation and significance.